**Clinical Concept and Relation Extraction Using Prompt-based Machine Reading Comprehension**

Authors:　　　　　　　　　Cheng Peng, PhD[1]
　　　　　　　　　　　　　Xi Yang, PhD[1,2]
　　　　　　　　　　　　　Zehao, Yu, MS[1]
　　　　　　　　　　　　　Jiang Bian, PhD[1,2]
　　　　　　　　　　　　　William R. Hogan, MD, MS[1]
　　　　　　　　　　　　　Yonghui Wu, PhD[1,2]

Affiliation of the authors:　[1]Department of Health Outcomes and Biomedical Informatics, College of Medicine, University of Florida, Gainesville, Florida, USA

　　　　　　　　　　　　　[2]Cancer Informatics Shared Resource, University of Florida Health Cancer Center, Gainesville, Florida, USA

Corresponding author:　　　Yonghui Wu, PhD
　　　　　　　　　　　　　Clinical and Translational Research Building
　　　　　　　　　　　　　2004 Mowry Road, PO Box 100177
　　　　　　　　　　　　　Gainesville, FL, USA, 32610
　　　　　　　　　　　　　Phone: 352-294-8436
　　　　　　　　　　　　　Email: yonghui.wu@ufl.edu

# ABSTRACT

**Objective**

To develop a natural language processing system that solves both clinical concept extraction and relation extraction in a unified prompt-based machine reading comprehension (MRC) architecture with good generalizability for cross-institution applications.

**Methods**

We formulate both clinical concept extraction and relation extraction using a unified prompt-based MRC architecture and explore state-of-the-art transformer models. We compare our MRC models with existing deep learning models for concept extraction and end-to-end relation extraction using two benchmark datasets developed by the 2018 National NLP Clinical Challenges (n2c2) challenge (medications and adverse drug events) and the 2022 n2c2 challenge (relations of social determinants of health [SDoH]). We also evaluate the transfer learning ability of the proposed MRC models in a cross-institution setting. We perform error analyses and examine how different prompting strategies affect the performance of MRC models.

**Results and Conclusion**

The proposed MRC models achieve state-of-the-art performance for clinical concept and relation extraction on the two benchmark datasets, outperforming previous non-MRC transformer models. GatorTron-MRC achieves the best strict and lenient F1-scores for concept extraction, outperforming previous deep learning models on the two datasets by 1%~3% and 0.7%~1.3%, respectively. For end-to-end relation extraction, GatorTron-MRC and BERT-MIMIC-MRC achieve the best F1-scores, outperforming previous deep learning models by 0.9%~2.4% and 10%-11%, respectively. For cross-institution evaluation, GatorTron-MRC outperforms traditional GatorTron by 6.4% and 16% for the two datasets, respectively. The proposed method is better at handling nested/overlapped concepts, extracting relations, and has good portability for cross-institute applications.

**INTRODUCTION**

Identifying clinical concepts (e.g., medications) and associated relations (e.g., medication-caused adverse events) is a fundamental natural language processing (NLP) task to facilitate the use of clinical text for health outcomes and translational studies [1]. Current computational algorithms in the medical domain heavily rely on structured patient information while important information remains locked in the narrative clinical text [2]. The clinical NLP community has developed corpora and NLP systems for clinical concept and relation extraction through a series of open challenges [3]. Various NLP methods including rule-based, machine learning-based and hybrid systems have been developed[4,5]. Recently, deep learning-based NLP models, especially transformer models, have remarkably improved the performance for clinical concept and relation extraction [6,7]. Nevertheless, there are still challenges, including (1) overlapped and nested concepts, e.g., one concept (or partial) can be annotated as more than one category, or have multiple relations [8,9]; (2) efficiently extract relations, most studies enumerate combinations among all concepts and then perform classification, which often causes serious imbalanced positive-negative ratio as only a small subset of the combinations have relations [3,8]; and (3) the portability of concept and relation extraction when applied to a cross-institution setting[10–12].

Breakthroughs in general NLP models provide opportunities to address these challenges. Recently, a machine reading comprehension (MRC) architecture shows a better ability to handle nested/overlapped concepts and the prompt-based learning algorithms show promising transfer learning ability to improve the portability of NLP models for cross-institution applications. This study approaches clinical concept and relation extraction using a prompt-based MRC architecture, where both concepts and relations can be identified by answering questions other than previous solutions based on sequence labeling and classification. We compared our methods with previous non-MRC deep learning-based solutions using two clinical benchmark datasets, including the 2018 National NLP Clinical Challenges (n2c2) challenge (focusing on medications and adverse drug events)[13] and the 2022 n2c2 challenge (focusing on social determinants of health [SDoH] and various attributes)[14]. We explored four pre-trained large transformer models including BERT, BERT-MIMIC, RoBERTa-MIMIC, and GatorTron[15], which is a large language model developed using over 90 billion words of text (with 82 billion words of clinical text). We also examined the portability of our MRC models in a cross-institution setting where the test set is from a different institution than the training. The experimental results show that our MRC models remarkably improved concept and relation extraction for medications and SDoH, and that our MRC models have better portability when applied to a cross-institution setting.

**BACKGROUND**

Clinical concept extraction (or named entity recognition [NER]) and relation extraction (RE) are two fundamental NLP tasks in the clinical domain[16–19]. Early studies of clinical concept extraction are often rule-based systems [20,21] that use human-defined rules and dictionaries to capture documentation patterns in clinical text. Later, supervised machine learning models such as the conditional random fields (CRFs)[22] and support vector machines (SVMs)[23] are applied. Recently, deep learning models, such as convolutional neural network (CNN)[24], recurrent neural network (RNN)[25] and its variants long short-term memory (LSTM)[26] further released human from feature engineering using a distributed representation. Specifically, Lample *et al.* [27] propose an RNN model implemented using LSTM architecture with a CRFs layer for clinical concept extraction; Yu *et al.* [28] demonstrate that the bidirectional LSTMs with a biaffine classifier are better at recognizing overlapped and nested concepts. These models utilize distributed word representations derived from large-scale unlabeled text using word embedding algorithms such as word2vec[29], GloVe[30], and FastText[31]. More recently, inspired by the self-attention mechanism[32], transformer architectures have become the mainstream solution for clinical NLP because of their better ability to handle long-term dependencies and high parallelization capabilities. Various transformer-based models such as BERT[33], ALBERT[34], RoBERTa[35], and ELECTRA[36] have been proposed and achieved state-of-the-art performance. Yang *et al.* [37] systematically explored four transformer architectures for clinical concept extraction.

Clinical RE is often approached as a classification task to determine relations among clinical concepts in a document. Typically, an end-to-end clinical RE system consists of a first step to identify concepts (i.e., clinical concept extraction) and a second step to classify relations. Early studies explored traditional machine learning models such as SVMs, random forests (RFs), and other kernel-based models and examined various context features (e.g., statistical linguistic features), knowledge features, and graph structures. For example, the best clinical RE system in the 2010 i2b2 challenge was developed using SVMs[38]; Chapman *et al.*[39] developed an RE system based on RFs and achieved the best performance in the 2018 MADE1.0 challenge; Li *et al.*[40] introduced neural networks to model the shortest dependency path and sentence sequence between target entities in the 2012 i2b2 challenge; Christopoulou *et al.*[41] developed an ensemble deep learning method for extraction of medications and adverse drug event (ADE) achieved the best micro-averaged F1-scores of 0.9472 and 0.8765 for relation-only and end-to-end settings, respectively. Recently, transformer-based NLP models have been successfully applied for clinical RE. Yang *et al.* [42] systematically examined the performance of three transformer models (i.e., BERT, RoBERTa, and XLNet) for relation extraction from clinical text based on a multi-class classification strategy and a binary classification strategy and demonstrated the effectiveness of clinical transformer models.

Most existing work approach clinical concept extraction as a sequence labeling task and adopted the classic 'BIO' tags to represent words at the beginning of a concept, inside of a concept, and outside of any concept. However, this mechanism cannot be used to naturally represent overlapped or nested concepts in a unified NLP system [43]. Ad hoc solutions have been proposed to alleviate this issue by adopting the "*one model per concept category*" strategy – training separate models for each individual concept. Most transformer-based clinical RE systems have also adopted the previous strategy of enumerating the combinations among concepts for classification [42], which is a time-consuming procedure that generates a large number of negative samples, causing an imbalanced distribution of positive-negative samples. In addition, information extraction systems based on these architectures often suffer from documentation variations when applied to datasets from different institutions and disease domains. NLP models that can better handle the documentation variations across institutions are needed.

Most recently, prompt-based learning architectures have demonstrated good few-shot learning abilities to better adapt to new samples that are not covered in the training data, which provides a great opportunity to improve the portability of clinical NLP systems across institutions[44]. Machine reading comprehension (MRC) has recently demonstrated better performance for concept and relation extraction with the ability to better handle overlapped or nested concepts [45]. MRC is an NLP task that aims to answer questions by extracting relevant information from a given context. Recent studies reported that MRC architecture have good concept extraction and relation extraction ability in the general domain. Levy *et al.* [46] showed that relation extraction can be approached by asking MRC models to answer simple questions instead of classifications. Li *et al.* [47] formulated the entity and relation extraction as a multi-turn question answering problem which can be solved using MRC to identify the answer using reinforcement learning. Li *et al.* [48] also proposed a new solution for flat and nested NER tasks based on the MRC framework to account for the limitations of previous methods based on the sequence labeling framework. This study seeks to explore this novel MRC architecture for clinical concept and relation extraction to better handle the challenges identified in traditional solutions based on sequence labeling and classification.

## MATERIALS AND METHODS

**Dataset**

This study used the 2018 n2c2 dataset (track 2) focusing on relations of medication and adverse drug events (referred to as the drug-ADE dataset), and the 2022 n2c2 dataset (track 2) focusing on social determinants of health (SDoH) and associated attributes (referred to as the SDoH dataset). The drug-ADE dataset [14] consists of 505 discharge summaries from the Medical Information Mart for Intensive Care (MIMIC)-III database with annotations of 9 categories of clinical concepts (drug, drug attributes, ADEs) and 8 types of relations among drugs, drug-associated attributes, and ADEs. The SDoH dataset consists of 5 categories of SDoH concepts and 9 categories of SDoH-associated attribute concepts, and 28 categories of relations among SDoH concepts and SDoH-associated attributes. The 2022 n2c2 challenge provides two datasets including the MIMIC datasets (including MIMIC-train and MIMIC-test, developed using notes from the MIMIC III database) and the University of Washington dataset (UW-test) developed using clinical notes from UW. **Table S1** in the supplementary material shows summary statistics of the two datasets.

**Formulate clinical concept and relation extraction as MRC**

We define a piece of clinical text as a sequence $X = \{x_1, x_2, \ldots, x_n\}$ and a triplet as $\{\pi = (e_1, r, e_2) | e_1 \in E_1, e_2 \in E_2, r \in R\}$, where $E_1$ and $E_2$ are pre-defined sets of clinical concept categories, $R$ is a predefined set of all relation categories, $n$ represents the sequence length and $x_i$ is the i-th word in the sequence. The goal of clinical concept and relation extraction is to recognize all triplets $\pi$ from a given $X$. In this definition, the triplets can share the same concepts or relations, i.e., the nested or overlapping entities. We approached clinical concept extraction using a prompt-based MRC architecture as shown in Figure 1, where concepts (i.e., $E$) were identified by asking concept-related questions (i.e., the prompts). Similarly, we approached relation extraction (i.e., the triplets) by asking MRC models to identify concepts that have a relation with the target concept using the relation-related questions as prompts. Thus, the end-to-end RE, i.e., the extraction of the triplet $(e_i, r_{ij}, e_j)$, can be solved using a unified MRC architecture as shown in Figure 2.

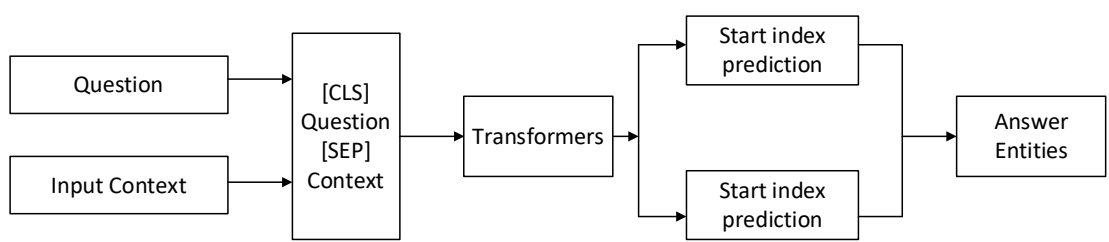

**Figure 1**. An overview of the machine reading comprehension (MRC) architecture. In MRC, the transformer models are used to identify the spans (i.e., the start index, and end index) from the text.

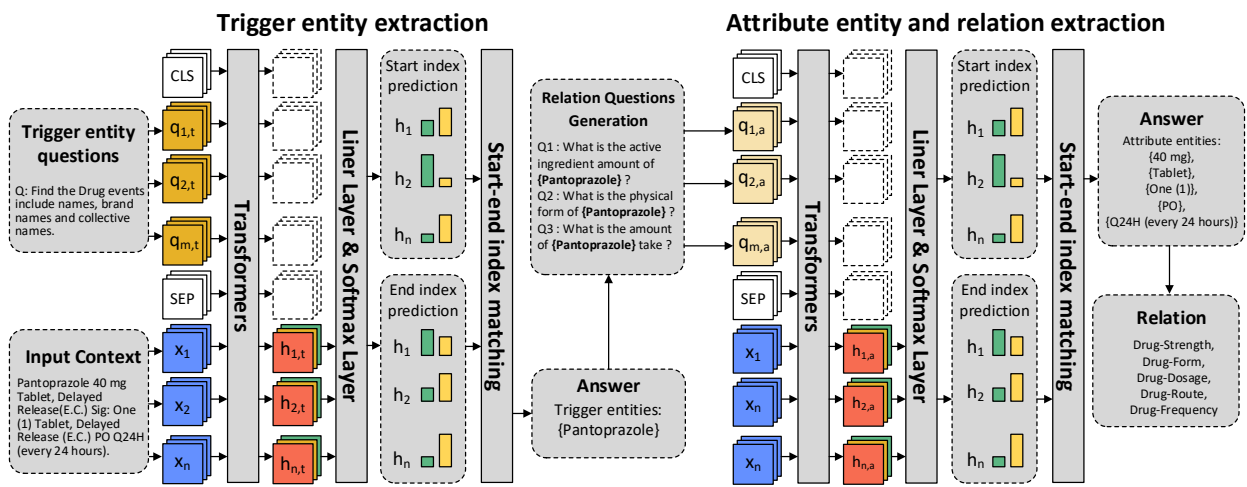

**Figure 2**. An overview of the unified machine reading comprehension (MRC) architecture for end-to-end clinical RE.

As shown in Figure 2, we first applied MRC to identify the trigger concept $e_i$ – defined as the concept serving as a hub to form relations with other concepts (e.g., "Drug" in the drug-ADE dataset, "Employment" in the SDoH dataset), and then generate the relation questions using these trigger concepts. For example, we used the following question, "find the drug events including names, brand names and collective names" as a prompt for MRC models to identify the drug mentions, which were then used to generate relation-related questions as new prompts to find other concepts $e_j$ - the "Strength" of drugs. The detailed model architecture and loss functions are provided in the supplement material.

**Question (or prompt) generation**

The MRC architecture identifies concepts and relations according to predefined questions, which capture the definitions of concepts and relations of interest (as the "prompts") to instruct MRC models identify the correct answer. The questions can be natural language questions (e.g., "What drug is mentioned in the text") or pseudo-questions (e.g., "entity: drug"). Following previous studies[47,48], we adopt the natural language questions that are constructed based on the annotation guidelines.

**Experiments and evaluation**

We explored four transformer models including BERT, BERT-MIMIC, RoBERTa-MIMIC, and GatorTron using the MRC architecture and compared them with existing transformer models using non-MRC architectures. Our previous studies[37,42] showed that BERT, BERT-MIMIC and RoBERTa-MIMIC are among the top-performing transformer models for clinical concept and relation extraction. GatorTron[15] is a new large language model pretrained using over 90 billion words of text and achieved state-of-the-art performance for five clinical NLP tasks. Specifically, for the 2018 n2c2 benchmark dataset, we compared our MRC models with previous models based on LSTM-CRFs, BERT, RoBERTa, ALBERT, and ELECTRA. For the 2022 n2c2 benchmark dataset, we compared our MRC models with the corresponding transformer models that were developed using non-MRC architectures. We also examined the generalizability of our MRC models when applied to cross-institution settings using the 2022 n2c2 MIMIC-train and the UW-test datasets.

BERT: a bidirectional transformer-based encoder language model pre-trained over a large general English domain corpus. BERT adopted the masked language modeling (MLM) and next-sentence prediction (NSP) training objectives to create deep representations capturing contextual information.

RoBERTa: a transformer-based model with the same architecture as BERT but pre-trained using a dynamic masked language modeling and optimized using different strategies (e.g., removing the next sentence prediction).

GatorTron[15]: a large clinical language model developed in our previous work, which is pre-trained from scratch using >90 billion words of text (including >82 billion words of de-identified clinical text). In our experiment, we used the GatorTron-base model which contains 345 million parameters for evaluation.

***Experiment settings:*** We developed our MRC-based models using existing transformer architectures implemented in the Transformers library developed by the HuggingFace team[49] using the PyTorch Lightning library. For general transformer models, we used the models pre-trained using general English text from the HuggingFace model repository. For the clinical transformer models, we used the models pre-trained using the MIMIC III corpus (BERT-MIMIC, RoBERTa-MIMIC, ALBERT-MIMIC and ELECTRA-MIMIC) in our previous study[37,42]. We adopted a five-fold cross-validation strategy to optimize hyperparameters, including the learning rate (1e-6, 1e-5, and 3e-5) and the training batch size (1, 4, and 8). The best models were selected according to the cross-validation performances measured by micro-averaged strict F1-score. All experiments were conducted using five Nvidia A100 GPUs. We reported the strict and lenient micro-averaged precision, recall, and F1-score aggregated from all concept and relation categories. The official evaluation scripts provided by the 2018 n2c2 challenges were used to calculate the scores.

**RESULTS**

Table S2 and S3 in the supplement material show the questions (prompts) we used for the 2018 n2c2 dataset and 2022 n2c2 dataset. Table 1 compares our MRC models with previously reported non-MRC models for clinical concept extraction. For the drug-ADE benchmark dataset, all four MRC models outperformed previous non-MRC models. Among the three MRC models, the GatorTron-MRC model achieved the best strict and lenient F1-scores of 0.9059 and 0.9506, respectively. GatorTron-MRC outperformed non-MRC models by 1%~3% in strict F1-scores. The other three MRC models BERT-MRC, BERT-MIMIC-MRC, and RoBERTa-MIMIC-MRC also achieved competitive performance with strict F1-scores of 0.9018, 0.9031 and 0.9020, respectively. Similarly, for the SDoH dataset, all four MRC models outperformed non-MRC models in strict F1-scores. Our GatorTron-MRC model achieved the best strict and lenient F1-scores of 0.8451 and 0.9342, respectively, outperforming non-MRC models by 0.7~1.3% in strict F1-scores.

**Table 1**. Comparison of MRC models with non-MRC transformer models for clinical concept extraction.

| Corpus | Models | | Strict Scores | | | Lenient Scores | | |
|---|---|---|---|---|---|---|---|---|
| | | | Precision | Recall | F1 | Precision | Recall | F1 |
| 2018 n2c2 | Non-MRC models | LSTM-CRFs[8] | 0.8893 | 0.8728 | 0.8810 | 0.9392 | 0.9184 | 0.9287 |
| | | BERT[37] | 0.8887 | 0.8728 | 0.8807 | 0.9398 | 0.9174 | 0.9285 |
| | | BERT-MIMIC[37] | 0.8835 | 0.8871 | 0.8853 | 0.9367 | 0.9349 | 0.9353 |
| | | RoBERTa[37] | 0.8821 | 0.8804 | 0.8812 | 0.9342 | 0.9276 | 0.9309 |
| | | RoBERTa-MIMIC[37] | 0.8927 | 0.8888 | 0.8907 | 0.9425 | 0.9333 | 0.9379 |
| | | ALBERT[37] | 0.8772 | 0.8766 | 0.8769 | 0.9311 | 0.9255 | 0.9283 |
| | | ALBERT-MIMIC[37] | 0.8776 | 0.8909 | 0.8842 | 0.9272 | 0.9363 | 0.9317 |
| | | ELECTRA[37] | 0.8689 | 0.8781 | 0.8735 | 0.9281 | 0.9314 | 0.9286 |
| | | ELECTRA-MIMIC[37] | 0.8814 | 0.8857 | 0.8836 | 0.9319 | 0.9387 | 0.9325 |
| | | BiLSTM-CRF[14] | 0.8973 | 0.8939 | 0.8956 | 0.9461 | 0.9376 | 0.9418 |
| | MRC models | **BERT-MRC** | 0.9159 | 0.8942 | 0.9018 | 0.9651 | 0.9202 | 0.9440 |
| | | **BERT-MIMIC-MRC** | 0.9187 | 0.8978 | 0.9031 | 0.9679 | 0.9221 | 0.9489 |
| | | **RoBERTa-MIMIC-MRC** | 0.9165 | 0.8964 | 0.9020 | 0.9658 | 0.9210 | 0.9465 |
| | | **GatorTron-MRC** | **0.9199** | **0.9012** | **0.9059** | **0.9708** | **0.9281** | **0.9506** |
| 2022 n2c2 | Non-MRC models | BERT | 0.8160 | 0.8483 | 0.8318 | 0.9122 | 0.9379 | 0.9249 |
| | | BERT-MIMIC | 0.8190 | 0.8512 | 0.8378 | 0.9193 | 0.9481 | 0.9320 |
| | | GatorTron | 0.8181 | 0.8508 | 0.8341 | 0.9171 | 0.9469 | 0.9318 |
| | MRC models | **BERT-MRC** | 0.8496 | 0.8353 | 0.8424 | 0.9328 | 0.9345 | 0.9325 |
| | | **BERT-MIMIC-MRC** | 0.8513 | 0.8390 | 0.8435 | 0.9398 | 0.9386 | 0.9337 |
| | | **RoBERTa-MIMIC-MRC** | 0.8486 | 0.8350 | 0.8412 | 0.9317 | 0.9338 | 0.9321 |
| | | **GatorTron-MRC** | **0.8521** | **0.8396** | **0.8451** | **0.9402** | **0.9398** | **0.9342** |

**Table 2**. Comparison of MRC models with non-MRC models for end-to-end clinical relation extraction.

| Corpus | Model | | Strict Criterion | | | Lenient Criterion | | |
|---|---|---|---|---|---|---|---|---|
| | | | Precision | Recall | F1 | Precision | Recall | F1 |
| 2018 n2c2 | Non-MRC models | LSTM-CRF+SVMs[8] | 0.8337 | 0.7773 | 0.8045 | 0.9112 | 0.8468 | 0.8778 |
| | | BiLSTM-CRFs[50] | 0.8536 | 0.7884 | 0.8197 | 0.9292 | 0.8549 | 0.8905 |
| | | BERT-MIMIC | 0.8492 | 0.7835 | 0.8141 | 0.9210 | 0.8512 | 0.8891 |
| | MRC models | **BERT-MRC** | 0.8582 | 0.7913 | 0.8251 | 0.9311 | 0.8553 | 0.8908 |
| | | **BERT-MIMIC-MRC** | 0.8601 | 0.7967 | 0.8279 | 0.9327 | 0.8587 | 0.8918 |
| | | **RoBERTa-MIMIC-MRC** | 0.8593 | 0.7932 | 0.8261 | 0.9315 | 0.8565 | 0.8910 |
| | | **GatorTron-MRC** | **0.8612** | **0.7998** | **0.8291** | **0.9341** | **0.8608** | **0.8921** |
| 2022 n2c2 | Non-MRC models | BERT | 0.6186 | 0.6465 | 0.6322 | 0.7716 | 0.7967 | 0.7839 |
| | | BERT-MIMIC | 0.6273 | 0.6519 | 0.6401 | 0.7829 | 0.8019 | 0.7921 |
| | | GatorTron | 0.6269 | 0.6498 | 0.6395 | 0.7812 | 0.7996 | 0.7913 |
| | MRC models | **BERT-MRC** | 0.7358 | 0.7227 | 0.7325 | 0.8639 | 0.8535 | 0.8626 |
| | | **BERT-MIMIC-MRC** | **0.7469** | **0.7319** | **0.7426** | **0.8725** | **0.8610** | **0.8712** |
| | | **RoBERTa-MIMIC-MRC** | 0.7420 | 0.7215 | 0.7358 | 0.8698 | 0.8529 | 0.8663 |
| | | **GatorTron-MRC** | 0.7398 | 0.7280 | 0.7364 | 0.8690 | 0.8578 | 0.8679 |

**Table 2** compares our MRC models with previously reported non-MRC transformer models for end-to-end clinical relation extraction. For the drug-ADE extraction, all four MRC models outperformed non-MRC models. Among the four MRC models, the GatorTron-MRC model achieved the best strict and lenient F1-scores of 0.8291 and 0.8921, respectively, outperforming non-MRC models by 0.9%~2.4% in strict F1-score. Similarly, for the end-to-end task of extracting SDoH and associated attributes, all MRC models outperformed non-MRC models. Among the four

MRC models, BERT-MIMIC-MRC achieved the best strict and lenient F1-scores of 0.7426 and 0.8712, respectively, outperforming non-MRC models by 10%~11% in strict F1-scores.

**Table 3.** Cross-institute comparison of GatorTron models applied in MRC architecture with applied in traditional architecture using the 2022 n2c2 SDoH dataset.

| Task | Training | Test | Model | Strict F1 |
|---|---|---|---|---|
| SDoH concepts and attributes extraction | MIMIC-train | MIMIC-test | GatorTron | 0.8341 |
| | | | GatorTron-MRC | **0.8452** |
| | MIMIC-train | UW-test | GatorTron | 0.7488 |
| | | UW-test | GatorTron-MRC | **0.8124** |
| End-to-end relation extraction | MIMIC-train | MIMIC-test | GatorTron | 0.6395 |
| | | MIMIC-test | GatorTron-MRC | **0.7364** |
| | MIMIC-train | UW-test | GatorTron | 0.5526 |
| | | UW-test | GatorTron-MRC | **0.7122** |

Table 3 compares GatorTron models applied in the MRC architecture (i.e., GatorTron-MRC) with those applied in traditional architecture (i.e., GatorTron) that based on sequence labeling and classification, using two evaluation settings including (1) training on MIMIC-train and testing on MIMIC-test, (2) training on MIMIC-train and testing on UW-test. When the test dataset is from the same data source as the training (MIMIC), GatorTron-MRC outperformed GatorTron by 1.1% and 9.6% in F1-scores for clinical concept extraction and end-to-end relation extraction, respectively. But when the test dataset is from a different institution (UW-test), GatorTron-MRC outperformed traditional GatorTron by 6.4% and 16% for the two tasks, respectively.

# DISCUSSION

Clinical concept and relation extraction are fundamental tasks to facilitate text analytics for clinical and translational studies. This study developed a novel clinical concept extraction and relation extraction method using a prompt-based MRC architecture and demonstrated its effectiveness using two clinical benchmark datasets. The experimental results show that our MRC models achieved state-of-the-art performance for extracting relations of drugs and ADEs as well as SDoH, outperforming existing non-MRC deep learning models. The performance improvement is bigger for end-to-end relation extraction of SDoH (10%~11% increase in F1-score) compared with drug-ADE (0.7~1.3% increase). One potential reason is that the drug-ADE dataset has remarkably less relation categories (8 relations) than the SDoH dataset (28 relations). Therefore, the imbalanced positive/negative ratio caused by the traditional solution is more serious in the SDoH dataset than the drug-ADE dataset. Our MRC models also demonstrated good generalizability in a cross-institute setting owning to the transfer learning ability inherited from the prompt learning-based MRC architecture. The performance of our GatorTron-MRC model only dropped 3.2% (from 0.8452 to 0.8124, for concept) and 2.4% (0.7364 to 0.7122, for end-to-end relation extraction) in F1-scores when applied to the cross-institute setting, whereas GatorTron with traditional deep learning architecture dropped 8.5% (0.8341 to 0.7488) and 8.7% (0.6395 to 0.5526), respectively.

We conducted an error analysis to examine how the proposed MRC architecture improves clinical concept and relation extraction. First, the MRC models perform better for overlapped or nested concepts and their relations. For example, in the 2018 n2c2 dataset, the phrase "as long as your rash is itching" is annotated as a "Duration", in which "rash" and "itching" respectively refer to the "ADE" and "Reason" concept overlapping with the "Duration" concept. Non-MRC models based on sequence labeling architecture have difficulties modeling these complicated cases using the traditional 'BIO' tag representations, where only one tag is allowed for a word. Our MRC architecture applies a question-answering architecture and implements two softmax classifiers to identify the span of answers, which provides a natural representation for these complicated cases that are often misclassified/missed in previous non-MRC models. In the MRC architecture, "rash" and "itching" are identified as answers for the ADE-specific and Reason-specific questions, respectively, and the phrase "as long as your rash is itching" is identified as the answer for the Duration-specific question. The proposed MRC architecture also extended previous MRC models that were developed using "one answer per question setting" by using two softmax classifiers to predict whether the input word is a start/end index. Similarly, this MRC architecture formulated RE as question answering to find concepts (*i.e.,* answers) for a given relation-specific question, which doesn't require enumeration over all combinations among all concepts for classification, which is time-consuming and often generates a large number of negative pairs, resulting in high computation cost and extremely imbalanced positive-negative ratio.

We further compare two strategies to generate questions from annotation guidelines including (1) natural language questions and (2) pseudo-questions. The relation-specific pseudo questions are often generated using a combination of head entity, relation type, and tail entity. Table 4 compares the two strategies using the 2022 n2c2 dataset. The experimental results show that natural language questions, which provide more fine-grained descriptions, are better than pseudo questions.

**Table 4**. Comparison of two question generation strategies using the 2022 n2c2 dataset.

| MRC model | Question generation strategy | Concept extraction (F1-score) | End-to-end relation extraction (F1-score) |
|---|---|---|---|
| BERT-MRC | Pseudo-questions | 0.8378 | 0.7256 |
|  | Natural language questions | **0.8424** | **0.7325** |
| BERT-MIMIC-MRC | Pseudo-questions | 0.8416 | 0.7298 |
|  | Natural language questions | **0.8435** | **0.7426** |
| GatorTron-MRC | Pseudo-questions | 0.8413 | 0.7252 |
|  | Natural language questions | **0.8451** | **0.7364** |

We limited the question (i.e., prompt) generation as a human engineering input (i.e., discrete/hard prompts), which rely on human expert with prior knowledge. Recent progress in prompt-based learning is exploring algorithms to train machines to learn continuous/soft prompts[51,52]. Due to the limited clinical resource, we evaluated the MRC architecture using two datasets focusing on the relations of medications with ADEs and SDoH. Our future work will investigate algorithms for learning soft prompts, explore strategies for handling cross-sentence relations, and examine the MRC architecture for other clinical NLP tasks not limited to concept and relation extraction.

**CONCLUSION**

This study presents and evaluates a unified MRC architecture for clinical concept and relation extraction. The proposed architecture provides a better solution to handle overlapped/nested clinical concepts and relations in a unified model and demonstrates better portability when applied to a cross-institution setting.

## ACKNOWLEDGMENTS

We would like to thank the i2b2 and n2c2 challenge organizers to provide the annotated corpus. We gratefully acknowledge the support of NVIDIA Corporation with the donation of the GPUs used for this research.

## FUNDING STATEMENT

This study was partially supported by a Patient-Centered Outcomes Research Institute® (PCORI®) Award (ME-2018C3-14754), a grant from the National Cancer Institute, 1R01CA246418 R01, a grant from the National Institute on Aging, NIA R21AG062884, and the Cancer Informatics and eHealth core jointly supported by the UF Health Cancer Center and the UF Clinical and Translational Science Institute. The content is solely the responsibility of the authors and does not necessarily represent the official views of the funding institutions.

## COMPETING INTERESTS STATEMENT

Cheng Peng, Xi Yang, Zehao Yu, Jiang Bian, William R. Hogan, and Yonghui Wu have no conflicts of interest that are directly relevant to the content of this study.

## CONTRIBUTORSHIP STATEMENT

CP, XY and YW were responsible for the overall design, development, and evaluation of this study. CP and ZY performed the experiments. CP and YW did the initial drafts of the manuscript, XY, ZY, JB, and WRH also contributed to writing and editing of this manuscript.

All authors reviewed the manuscript critically for scientific content, and all authors gave final approval of the manuscript for publication.

## SUPPLEMENTARY MATERIAL

Attached in a separate document.

**Transformer-based machine reading comprehension (MRC) architecture**

We applied transformer-based models to solve MRC. Specifically, the goal is to identify the span $x_{\text{start,end}}$ of an answer from a given text $X = \{x_1, x_2, \ldots, x_n\}$ based on a given question $Q = \{q_1, q_2, \ldots, q_m\}$, where $n$ and $m$ denote the number of words in X and Q, respectively. Taking BERT as an example, we concatenated the question Q with the given text X as the input $S = \{[\text{CLS}], q_1, q_2, \ldots, q_m, [\text{SEP}], x_1, x_2, \ldots, x_n\}$, where [CLS] indicates the start token of Q and [SEP] is special token defined in BERT to separate Q and X. Then, the input sequence is tokenized to a token sequence $s = [s_i]_{i=1}^{k}$ concatenating with their position embedding and segment embedding. The BERT encoder that consists of $L$ stacks of transformers as $BERT\,(\cdot)$ is formulated as follows:

$$h_i^l = Transformers(s_i^{l-1}), l \in [1, L] \tag{1}$$

$$H = [h_1^L, h_2^L, \ldots, h_n^L] \in R^{n \times d} \tag{2}$$

Where $h_i^l$ denotes the hidden representation of the input context in *l*-th layer, $d$ is the vector dimension of the last layer of BERT.

**Answer span identification**

We adopted two binary classifiers: one classifier to predict whether a token is the start index, and the other to predict whether a token is the end index. This strategy enabled us to identify multiple start indexes and multiple end indexes for a given input text. Specifically, given the representation matrix $H$ (defined by equation 2), the classifier performs a binary classification to calculate the probability of each token being a start index or an end index as follows:

$$P_{start} = \text{softmax}(\text{Linear}(H \cdot W_{start}) \cdot V_{start}) \in R^{n \times 2} \quad (3)$$

$$P_{end} = \text{softmax}(\text{Linear}(H \cdot W_{end}) \cdot V_{end}) \in R^{n \times 2} \quad (4)$$

Where, $W$ and $V$ are model parameters. If the input text contains multiple answers, there will be multiple start and end indexes identified, thus, to account for the overlapped and nested concepts. We trained a classifier to match the start index to its corresponding end index when there were multiple answers. Detailed descriptions of our algorithm are provided in the Supplement.

**Loss function**

We used cross entropy (CE) loss to optimize all the classifiers. Let $Y_{start}$ and $Y_{end}$ be the gold standard annotation, the loss function for the start and end index prediction is defined as:

$$L_{start} = CE(P_{start}, Y_{start}) \quad (5)$$

$$L_{end} = CE(P_{end}, Y_{end}) \quad (6)$$

Let $Y_{start,end}$ denote the gold standard labels indicating whether a start index should be matched with a given end index, we used the following loss function to train the classifier to match start and end index:

$$L_{span} = CE(P_{start,end}, Y_{start,end}) \quad (7)$$

Then, the overall loss function is formulated as follows:

$$L = \alpha L_{start} + \beta L_{end} + \gamma L_{span} \quad (8)$$

Where, $\alpha, \beta, \gamma \in [0,1]$ are hyper-parameters to control the contributions from each step. Therefore, we optimized all classifiers in a unified end-to-end model, where parameters were shared at the transformer models.

**Methods to match the start index to the corresponding end index for multiple answers**

By applying argmax to each row of $P_{start}$ and $P_{end}$, we can get the predicted indexes that might be the starting or ending positions, i.e., $I_{start}$ and $I_{end}$:

$$I_{start} = \{i | \text{argmax}(P^i_{start}) = 1, i = 1, 2, \ldots, n\} \tag{9}$$

$$I_{end} = \{i | \text{argmax}(P^i_{end}) = 1, i = 1, 2, \ldots, n\} \tag{10}$$

where $i$ denotes the $i$-th row of a matrix. Given any start index $i_{start} \in I_{start}$ and end index $i_{end} \in I_{end}$, a binary classification model is used to predict the probability that they should be matched, given as follows:

$$P_{i_{start,end}} = \text{sigmoid}\left(m \cdot \text{concat}(E_{i_{start}}, E_{j_{end}})\right) \tag{11}$$

where $m \in R^{1 \times 2d}$ is the weights to learn.

**Data preprocessing for MRC**

We converted the BRAT format data to MRC format instances (QUESTION, CONTEXT, ANSWER) using sentence level sample construction, i.e., extracting every single sentence from the input clinical text to construct the MRC instances. We created questions for all trigger concept categories ("Drug" in 2018 n2c2 dataset, "Employment", "Living status", "Alcohol", "Drug" and "tobacco" in 2022 n2c2 dataset) for each sentence. The MRC model identified answers as a span set if the sentence contains trigger concepts; otherwise, the answer is set to None. The identified trigger concepts were used to generate relation-specific questions.

Table S1. Summary statistics of the clinical notes and annotated concepts and relations in 2018 n2c2 drug-ADE dataset and the 2022 n2c2 SDoH datasets.

| Challenge | Datasets | Number of notes | Number of clinical concepts | Number of clinical relations |
| --- | --- | --- | --- | --- |
| 2018 n2c2 (Medication-ADE) | Training | 303 | 50951 | 36384 |
| | Test | 202 | 32918 | 23462 |
| 2022 n2c2 (SDoH, attributes) | Training (MIMIC) | 1316 | 16039 | 10933 |
| | Development (MIMIC) | 188 | 1744 | 1177 |
| | Test (MIMIC) | 373 | 3331 | 2243 |
| | UW-test | 518 | 4903 | 3249 |

ADE: adverse drug event; SDoH: social determinants of health: UW: University of Washington; MIMIC: Medical Information Mart for Intensive Care.

Table S2. Natural language questions (prompts) used for clinical concept extraction.

| Concept | Natural Language Question for MRC models |
| --- | --- |
| **2018 n2c2 - drug and adverse drug events** | |
| Strength | What is the active ingredient amount of Pantoprazole? |
| Form | What is the physical form of Pantoprazole? |
| Dosage | What is the amount of Pantoprazole taken? |
| Frequency | How often each dose of Pantoprazole should be taken? |
| Route | What is the path of Pantoprazole taken into the body? |
| Duration | How long to take Pantoprazole? |
| Reason | What is the medical reason for giving Pantoprazole? |
| ADE | What are the injuries resulting from the use of Pantoprazole? |
| **2022 n2c2 - social determinants of health and attributes** | |
| Employment | Find the employment event in the text, including work-related key phrases and subheadings. |
| Living status | Find the living status events that are form of "lives" or "resides". |
| Alcohol | Find the alcohol events like "alcohol", "ETOH", "drink", or "beer". |
| Drug | Find the drug events involve marijuana, illegal drugs, or the abuse of prescription drugs. |
| Tobacco | Find the tobacco events that are first-hand smoking or smokes. |
| TypeLiving | Find the type of living like alone, with family, with others, or homeless. |
| StatusEmploy | Find the status of employment like employed, unemployed and retired. |
| Method | How the substance is used. |
| Duration | How long has the substance use persisted like for years or over years. |
| Frequency | How often the substance is used like per or every or few times. |
| StatusTime | Find the status of substance use like none, current, or past. |
| Type | Find the specific type of substance used. |
| Amount | Find the amount of substance use. |
| History | How long ago the substance use occurred like years ago or in years. |

**Table S3**. Natural language questions (prompts) used for end-to-end relation extraction.

| Concept/relation | Natural Language Question for MRC models |
|:---|:---|
| **2018 n2c2 drug and adverse drug events** | |
| **Trigger concept questions** | Find the drug events including names, brand names and collective names. |
| **Relation Questions** | |
| Strength | What is the active ingredient amount of Pantoprazole? |
| Form | What is the physical form of Pantoprazole? |
| Dosage | What is the amount of Pantoprazole taken? |
| Frequency | How often each dose of Pantoprazole should be taken? |

|  | Route | What is the path of Pantoprazole taken into the body? |
|---|---|---|
|  | Duration | How long to take Pantoprazole? |
|  | Reason | What is the medical reason for giving Pantoprazole? |
|  | ADE | What are the injuries resulting from the use of Pantoprazole? |
| **2022 n2c2 social determinants of health and attributes** | | |
| **Trigger concept questions** | | |
|  | Employment | Find the employment event in the text, including work-related key phrases and subheadings. |
|  | Living status | Find the living status events that are form of "lives" or "resides". |
|  | Alcohol | Find the alcohol events like "alcohol", "ETOH", "drink", or "beer". |
|  | Drug | Find the drug events involve marijuana, illegal drugs, or the abuse of prescription drugs. |
|  | Tobacco | Find the tobacco events that are first-hand smoking or smokes. |
| **Relation questions** | | |
| Employment | StatusEmploy | what is the status of employment associated with "Retired surgical nurse" |
|  | Type | what is the type of employment associated with "Retired surgical nurse" |
|  | Duration | how long has the employment associated with "Retired surgical nurse" lasted |
|  | History | how long ago the employment associated with "Retired surgical nurse" occurred |
| LivingStatus | Type | what is the type of living status associated with " lives " |
|  | StatusTime | what is the status of living associated with " lives " |
|  | Duration | how long the living status associated with " lives " lasted |
|  | History | how long ago the living status associated with " lives " occurred |
| Tobacco | Duration | how long the tobacco use associated with " nonsmoker " lasted |
|  | History | how long ago the tobacco use associated with " nonsmoker " occurred |
|  | StatusTime | what is the status of tobacco use associated with " nonsmoker " |
|  | Type | what is the specific type of tobacco use associated with " nonsmoker " |
|  | Amount | what is the amount of tobacco use associated with " nonsmoker " |
|  | Method | how the tobacco associated with " nonsmoker " is used |
|  | Frequency | how often the tobacco associated with " nonsmoker " is used |
| Alcohol | Duration | how long the alcohol use associated with " ETOH " lasted |
|  | History | how long ago the alcohol use associated with " ETOH " occurred |
|  | StatusTime | what is the status of alcohol use associated with " ETOH " |
|  | Type | what is the specific type of alcohol use associated with " ETOH " |
|  | Amount | what is the amount of alcohol use associated with " ETOH " |
|  | Method | how the alcohol associated with " ETOH " is used |
|  | Frequency | how often the alcohol associated with " ETOH " is used |
| Drug | Duration | how long the drug use associated with " IVDU " lasted |
| | History | how long ago the drug use associated with " IVDU " occurred |
| | StatusTime | what is the status of alcohol use associated with " IVDU " |
| | Type | what is the specific type of drug use associated with " IVDU " |
| | Amount | what is the amount of drug use associated with " IVDU " |
| | Method | how the drug associated with " IVDU " is used |
| | Frequency | how often the drug associated with " IVDU " is used |